# EVOLUTIONARY SEARCH IN THE SPACE OF RULES FOR CREATION OF NEW TWO-PLAYER BOARD GAMES

Zahid Halim
*Faculty of Computer Science and Engineering,*
*Ghulam Ishaq Khan Institute of Engineering Sciences and Technology, Topi, Pakistan*
*zahid.halim@giki.edu.pk*

Rauf Baig
*Al Imam Mohammad Ibn Saud Islamic University (IMSIU), Riyadh, Saudi Arabia*
*rauf.baig@nu.edu.pk*

Kashif Zafar
*Department of Computer Science,*
*National University of Computer and Emerging Science, Lahore, Pakistan*
*kashif.zafar@nu.edu.pk*

Games have always been a popular test bed for artificial intelligence techniques. Game developers are always in constant search for techniques that can automatically create computer games minimizing the developer's task. In this work we present an evolutionary strategy based solution towards the automatic generation of two player board games. To guide the evolutionary process towards games, which are entertaining, we propose a set of metrics. These metrics are based upon different theories of entertainment in computer games. This work also compares the entertainment value of the evolved games with the existing popular board based games. Further to verify the entertainment value of the evolved games with the entertainment value of the human user a human user survey is conducted. In addition to the user survey we check the learnability of the evolved games using an artificial neural network based controller. The proposed metrics and the evolutionary process can be employed for generating new and entertaining board games, provided an initial search space is given to the evolutionary algorithm.

## 1. Introduction

Games have an importance in everyone's life. Some of them are centuries old and others are a product of the last few decades. They have proven to be an excellent test bed for testing algorithms falling under the umbrella of computational intelligence. Researchers have developed software agents which not only play games but also have the ability of learning first the concept of the game itself. Games popular with the computational intelligence researchers can be categorized in many ways. One such categorization can be (a) classical games which are played indoors (e.g. chess and bridge) [1], (b) video games which need a computing resource and a video display [2], (c) robotic games which are replicas of popular games played by robots [3], and (d) physically activating games which are played on playgrounds requiring electronic gadgetry (e.g. electronic tiles, laser beams) [4].

Perhaps many of us have wondered how a popular game, e.g. chess, might have originated and what were its rules at the beginning and how and why they evolved over centuries to reach their present state. Also, can we introduce our own modifications to make the game even better? Video games, being a recent phenomenon, can have their history more easily traced. The popular ones have their versions upgraded, based on user surveys which act as a sort of fitness function for guiding their evolution. Everyone would agree that it would be nice to have a reliable method for game creation. The next question is whether an evolutionary algorithm can be applied for that purpose. The answer is yes, and initial ground breaking experiments have already been reported [5, 6]. We need a space of game rules in which the search for an optimum rule set can be made and a fitness function to guide the search. One way to specify the search space can be to generalize the rules of existing games of the same genre which we are trying to create. The design of fitness function is a much more difficult task.

For developing a fitness function, the fundamental question is: how do we compare two games of the same genre? The one providing more entertainment would be better. However, it is not simple to quantify entertainment. A source of entertainment for one person may be source of annoyance for another. Even though entertainment is subjective, some games are enjoyed by a vast majority of people. The common elements in popular games can be studied and attempts can be made to develop metrics for quantifying entertainment. These metrics can be used as a fitness function to guide the evolution for creating, new games.

In our present work, we report experiments in creation of new board games for two players. The novelty of this paper is that we employ evolutionary computation to address two issues in game development; a) measuring entertainment value of a board based game and b) automatic generation of entertaining board games. We propose an entertainment metric to quantitatively measure the entertainment value of the game. Our proposed metrics consider four criterion of entertainment, namely: (a) duration of the game, (b) intelligence required for playing the game, (c) dynamism exhibited by the pieces and (d) usability of the play area. We have further shown the utility of the proposed entertainment metrics by generating new and entertaining games through an evolutionary process which utilizes our entertainment metrics as fitness function, guiding the evolution towards an entertaining set of games. We define an initial search space for the evolution of new board based games. The search space is specified by taking some aspects of two popular board based games, chess and checkers. This search space is utilized to initialize and produce games using an evolutionary algorithm. In order to guide the evolutionary algorithm towards a more entertaining game, we develop a set of metrics for quantifying entertainment value of board games and use them as a fitness function. Further we create two different types of agents that play and evaluate the new games generated, against the proposed entertainment metrics. The first type of agent makes random but legal moves to play the game whereas the second agent is relatively intelligent as it uses min-max algorithm with a rule based evaluation function. We conduct a human user survey to counter check the entertainment value of the evolved games against the human's entertainment value. In addition to the user survey, the entertainment value of evolved games is also verified by learnibility of the evolved games using an artificial neural network based controller, as proposed in Schmidhuber's theory of artificial curiosity [7].

The paper is arranged as follows: section II lists the search space in detail; section III covers the theoretical studies on different entertainment theories and based upon these theories defines our proposed entertainment metrics along with the evolutionary algorithm used; section IV describes the software agent and its types which we use to evaluate the generated games; section V contains the detailed experimentations including the effect of different types of controllers, user survey and controller learnability and section VI concludes the paper.

## 2. Search Space

We first need to define a search space in which search for an entertaining set of rules can take place. One way to do this is to study the rules of some famous games and generalize them. Two famous board games, checkers and chess are selected for this purpose. Following is a discussion on different features of both these games and which features we select for our search space.

**2.1. Size of play area:** Both, the games checkers as well as chess, consist of a grid of 8x8 squares, alternating white and black. White squares are never used in checkers but in case of chess both are used. The size of play area in our search space is also a grid of 8x8 squares, alternating white and black and all squares can be used. In principle, this should have been one of the dimensions of search and one of the two possibilities (white squares never used, as is done in checkers or all the squares used, as is the case for chess) should be committed for a proposed game. However, if the white squares never used option is selected, then the search space collapses considerably. Therefore we opt for using both white and black squares.

**2.2. Types of pieces, their quantities & their initial positions**: Checkers has a total of 12 pieces, called checkers; these can later be converted to a king. King is not present initially but comes into being by promotion when it reaches the first row of the opponent. Number of kings is a variable quantity but maximum can be 12. Initial positions of the pieces are the three rows nearest to the player and are placed only on black cells. Chess consist of a total of six types of pieces which are: (1) piece king, (one piece) queen, (2) pieces knight, (2) pieces bishop, (2) pieces rook, (8) pieces soldiers. Initial positions of the pieces are the two rows nearest to the player.

Combining the rule space of chess and checkers we have total six types of pieces in our search space. Each type can have a minimum of 0 and a maximum of 16 pieces. However, total pieces should not be zero nor exceed 16. The initial positions of the pieces are the nearest three rows of a player. A cell can have one piece of type 0 to 6, type 0 means no piece is present in that cell. For the ease of reference we do not name the types rather we call them by their type number. Evolution decides the initial positions, the number of types present and the number of pieces of a type.

**2.3. Movement direction of each piece & step size:** The pieces of checkers can move only in diagonal, forward direction with a step size of one. The diagonal jumps are allowed if the cell behind the opponent's cell is empty. In case a sequence of jumps is possible then all of them have to be taken. Only king is allowed to jump in diagonal forward and backwards, but the step size is same as that of a normal (checker) piece. Turn passing is not allowed. In the game of chess each type has a different movement logic and step size. A king is allowed to move in all directions with a step size of one. Queen can also move in all directions but there is no limit on number of steps. A knight can also move in all directions, but the move should be L shaped, thus the step size becomes two in one direction and then one step either left or right. A jump over other pieces is allowed to a knight. Bishop can move only diagonally, forward and backwards, with no limit on steps. A rook moves straight, forward and backwards, with no limit on steps. Soldiers are allowed to move straight, with step size of one, except for opening soldier movement. Turn passing is not allowed in chess and only one move per turn is allowed.

The search space to evolve new games consists of the six movement logics found in chess. The step size for a piece can either be one or up to an occupied cell.

**2.4. Capturing logic:** In checkers capturing is done by jumping over an opponent's piece. Whereas in chess capturing is done by moving into the cell occupied by the opponent's piece. Jumping over pieces is not allowed in chess, except for knight. For the knight, nothing happens to the pieces being jumped over. In both cases (of chess and checkers) the opponent's piece dies and is removed from the board.

The search space that we define for the new games is as follows: for each type capturing is done by jumping over or moving into the opponent's cell. The result of capturing is death of captured piece. There must be a vacant cell behind the cells when jump over is performed. In case several jumps are possible in a sequence, then all of them will have to be taken.

**2.5. Game ending logic & rules defining wins & draws:** The game of checkers ends when no moves are possible for a player. The player with greater number of checkers is winner. If, both the players have same number of checkers then a draw is declared. In the game of chess a game ends if no move is possible for the king. The player whose king cannot move is loser. A game can be declared a draw by consent of both players.

In our search space we specify the type of piece for which the game ends if there are no more pieces left of that type. We call this piece a piece of honor. There can be zero or one piece type declared to be a piece of honor. The restriction of one type has been done to keep the search space limited. The evolution will only decide which type of piece has this honor. A game ends if a piece of honor of any player is dead or the player without any possible move is the loser. We restrict the maximum number of moves in a game to be a maximum of 100. If the game ends due to number of moves increasing 100 then a player with greater number of pieces is winner. If both have same number than a draw is declared.

**2.6. Conversion to pieces of some other type after reaching last row:** In checkers a piece is converted into a king when it reaches the last row (or the first row of opponent). In chess this option is available to soldiers only. They convert into a queen or any piece of choice.

   In our case evolution decides which type is convertible. Each piece has a conversion logic which decides which type it will convert to when last row is reached. There is also a chance that no type conversion occurs.

**2.7 If a piece is being capture, is it mandatory to capture it***?* In checkers it is mandatory but in chess it is not to kill/capture an opponent's piece if it's possible to do so. In our search space we keep both the options.

**2.8. Turn passing allowed***?* Both in chess and checkers turn passing is not allowed so we fix the same in our search space. Table 1 summarizes the search space.

Table 1. Summary of the Search Space for Evolving New Board Based Games

| Search Space Dimension | Selected Values |
| --- | --- |
| Play Area | Both white & black squares are used |
| Types of Pieces | 6 |
| Number of pieces/type | variable but at maximum 24 |
| Initial position | Both white & black squares of first 3 rows |
| Movement direction | All directions, straight forward, straight forward and backward, L shaped, diagonal forward |
| Step Size | One Step, Multiple Steps |
| Capturing Logic | Step over, step into |
| Game ending logic | No moves possible for a player,  no moves possible for the king |
| Conversion Logic | Depends upon rules of the game |
| Mandatory killed | Depends upon rules of the game |
| Turn passing allowed | No |

**2.9 Structure of the chromosome**

   Based upon the above search space, the structure of the chromosome used is listed in Fig 1. The chromosome consists of a total 50 genes. First 24 genes may contain values from 0 to 6 were 1 represents a piece of type 1, 2 for piece of type 2 and so on. Zero is interpreted as no piece. The piece type represented by gene 1 is placed in the cell 1 of the game board; piece type represented by gene 2 is placed in the cell 2 of the game board and so on.

| Gene | Title | Value |
|------|-------|-------|
| 1-24 | Placement of gene of each type | 0-6 |
| 25-30 | Movement logic of each type | 1-6 |
| 31-36 | Step Size | 0/1 |
| 37-42 | Capturing logic move into cell or jump over 0/1 | 0/1 |
| 43 | Piece of honor | 0-6 |
| 44-49 | Conversion Logic 0-6 | 0-6 |
| 50 | Mandatory to capture or not | 0/1 |

Fig. 1. Structure of the chromosome

Gene 25 to 30 represents movement logic for each piece type respectively, where 1 is for diagonal forward, 2 for diagonal forward and backward, 3 for all directions, 4 for L shaped movement, 5 for straight forward and backward and 6 for straight forward. Genes 31 to 36 are used for step size of each type, where 0 is used to indicate single step size and 1 for multiple step size. Genes 37 to 42 are used for step size of each type, where 0 is used to indicate step into and 1 for step over. Gene number 43 indicates the type of piece that will be the piece of honor, possible values include 0-6, where 1-6 indicate the piece type and 0 represents that there is no piece of honor in the game. Genes 44-49 represents the conversion logic, of piece type 1 to 6 respectively, when they reach the last row of the game board. Where 0 represents the piece will not be converted to any type and 1-6 represents the type of piece. The last gene represent whether if it is mandatory in the game to capture the opponent piece in case it could be, 0 represents no and 1 represents yes.

## 3. Fitness Function and Evolutionary Algorithm

Evolutionary algorithms need a mechanism to evaluate the fitness of individuals of the population. Our aim is to evolve new games which are entertaining to play. Hence the fitness function must have features which can somehow quantify the entertainment value of a game. Actually we are looking for an answer to the question: "What are the factors which makes one board game more interesting than another?" To answer this question a peek at the literature on psychology might help. Some theories which are of interest are described in the following paragraphs.

### 3.1. Theories on Entertainment

According to Csikszentmihalyi's theory of flow [8, 9], the optimal experience for a person is when he is in a state of flow. In this state the person is fully concentrated on the task that he is performing and has a sense of full control. The state of flow can only be reached if the task is neither too easy nor too hard. In other words the task should pose the right amount of challenge.

In addition to the right amount of challenge, Malone [10] proposes two more factors that make games engaging: fantasy and curiosity. If a game has the capability of evoking the player's fantasy and makes him feel that he is somewhere else or doing something exotic then that game is more enjoyable than a game which does not do so. Curiosity refers to the game environment. The game environment should have the right amount of informational complexity: novel but not incomprehensible. Koster's theory of fun [11] states that the main source of enjoyment while playing a game is the act of mastering it. If a game is such that it is mastered easily and the player does not learn anything new while playing then the enjoyment value of that game is low.

Rauterberg [12, 13] has introduced the concept of *incongruity* as a measure of interest in a task. Given a task, humans make an internal mental model about its complexity. Incongruity refers to the difference between the actual complexity of the task and the mental model of that complexity. We have positive congruity if this difference is positive and negative congruity otherwise. In case of negative incongruity a person would be able to accomplish the task easily. Interest in a task is highest when the incongruity is neither too positive nor negative. In case of large positive incongruity the humans have a tendency to avoid the task and in situations of large negative incongruity they get bored. This requirement of right amount of incongruity is similar to the right amount of challenge in the concept of flow mentioned above. It has been further proposed that in case of reasonable positive incongruity the humans have a tendency to learn more about the task so that their mental model comes at par with the actual complexity of the task.

Several other related and derivative works are available on this topic. Many of them are covered in Yananakakis's recent survey [14].

### 3.2. Quantification of Entertainment

In addition to the debate about the comprehensiveness or accuracy of the above mentioned theories, we are faced with the problem of converting them into something quantifiable for the problem at hand, i.e. two player board games. The three possible methods of collecting data are as follows:

(a) One obvious method of quantification can be to have a human player survey about a set of games. If the number of participants of the survey is large enough then some meaningful information about the games being compared can be obtained.

(b) Statistics obtained from the games played can be used.

(c) Physiological signals of the human players during the games can be used.

If we are using an evolutionary algorithm for evolving games then the number of game evaluations is usually in order of thousands. There is a population of games in each of the several generations which have to be evaluated and compared. This large magnitude excludes the reliance on human game playing and we cannot use option (a) or (c). The only option available is to collect statistics from automatic game playing.

Iida [15], in 2003, has proposed a measure of entertainment for games and used it to analyze the evolution of game of chess over the centuries. This measure is considered to be the pioneer in quantification of entertainment. Even though Iida's work is limited to chess variants the measure of entertainment can be easily applied to other board games. According to this measure, the entertainment value of a game is equal to the length of the game divided by the average number of moves considered by a player on his turn. The game is more entertaining if the value of this measure is low. The main idea is that the player should have many choices (moves) on the average and the length of the game should not be large. Long games with few choices per move are boring. The authors differentiate between possible moves and the moves considered by a player. The set of considered moves is smaller than the set of possible moves and the metric is based on the moves considered by a player. Except for Iida's work, all other research has been in context of computer (video) games and physically interactive games.

In [16] the authors introduce the uncertainty of game outcome as a metric of entertainment. If the outcome is known at an early stage then there is not much interest in playing it. Similarly if it is found at the last move then it is probably probabilistic. The outcome should be unknown for a large duration of the game and should become known in the last few moves.

Togelius [6] has presented an approach to evolve entertaining car racing tracks for a video game. Tracks were represented as b-splines and the fitness of a track depended on how an evolved neural network based controller (modeled after a player) performed on the track. The objectives were for the car to have made maximum progress in a limited number of time steps (high average speed), high maximum speed, and high variability in performance between trials. In [17] Togelius evolves entertaining computer games and uses a fitness function based on the learnability of the games. A game which is either too hard or too easy to learn (by a population of ANNs) is assigned a low fitness. All other games are assigned fitness according to the degree to which they can be learnt in a fixed time.

In [18] three metrics (which are combined into one) have been proposed for measuring the entertainment value of predator/prey games. The first metric is called appropriate level of challenge (T). It is calculated as the difference between the maximum of a player's lifetime and his average lifetime over N games. This metric has a higher value if the game is neither too hard nor too easy and the opponents are able to kill the player in some of the games but not always. The second metric is behavior diversity metric (S). It is standard deviation of a player's lifetime over N games. It has a high value if there is diversity in opponent's behavior. The third metric is spatial diversity metric $E\{H_n\}$. It is the average entropy of grid-cell visits by the opponents over N games. Its value is high if the opponents move all the time and cover the cells uniformly. This movement portrays aggressive opponent behavior and gives an impression of intelligent game play. The three metrics are combined into one single metric $I = [\gamma T + \delta S + \varepsilon E\{H_n\}]/[\gamma + \delta + \varepsilon]$ where I is the interest value of the predator/prey game; γ, δ and ε are weight parameters. Work in [31, 32] presents evolution of rules for predator/pray games based on somewhat same metrics as proposed in [18].

In [19], the authors have developed a computer game called "Glove" with three levels of incongruity: hard, easy and balanced. There assumption is that the player would get frustrated or bored respectively, with the first two settings and would enjoy with the third one. They argue that the actual complexity of a game can be defined as its difficulty level and the incongruity, i.e. the difference between the actual complexity and a player's mental complexity of a game can be measured indirectly by observing the player's behavior in the game. If the player progresses easily through a game or does not progress at all then the incongruity is too easy or hard respectively. If the player has just enough health to reach a game's goal, but not more than the incongruity is balanced.

An interesting fact is that data pertinent to interestingness of games can be obtained by measuring the physical signals of human beings. However, that again requires a tremendous number of games to be played and thus cannot be used for two player board games. Variations in physiological signals of game players have also been shown to correlate with the enjoyment experienced by them. Some signals which have been investigated include cardiovascular measures, skin responses, facial and jaw electromyography, and respiration. Most of these studies are related to computer games [9, 20-28, 33, 35, 36] or physically interactive games [29] and none were done in the context of board games except [21].

### 3.3. Proposed Fitness Function

In context of two person board games the opponent is part of the gaming environment and his capabilities have a direct influence on the entertainment value of a game. Since we are concerned with evolving games we assume an opponent of abilities equal to the player. By doing so, we ignore the effect of opponent while analyzing the above mentioned theories.

According to theory of flow, the gaming environment should pose a right amount of challenge to the player. It implies that the game which is able to take a player into the state of flow would be better than a game which is not able to do so. The first of Malone's three factors, i.e. challenge is the same as that covered by theory of flow. Furthermore, Rauterberg's theory of incongruity is also similar to the theory of flow. Koster's theory of fun and the third Malone's factor of curiosity are similar in nature. The state space of the game should be sufficiently rich so that even the experienced players would find fresh avenues to explore and new situations to master. Malone's second factor of fantasy is different from the rest and would relate to the interpretation of the board layout and the pieces (e.g. king and queen of chess, property buying of Monopoly).

Togelius has categorized the entertainment theories into two types. Static theories, which rely on monitoring the process of game playing, and dynamic theories, which require monitoring of the process of learning the game. Theory of Flow, Malone's factors and Rauterberg's theory can be considered as static theories. They can be used to judge the entertainment value of a game for a player given his current capabilities of playing that game (assuming the opponent player to be part of the game environment). Koster's theory is a dynamic theory and judges a game by observing how a player adapts to it over time.

In the present work we have based our fitness function on metrics which can be described as adhering to the static theories. However, after evolution we do check some of the evolved games for their compatibility with the dynamic theories.

We now turn our attention to what can be quantitatively measured in two-player board games. We can measure the average game duration, the average branching factor, the number of wins for the first and second player, the number of games drawn, the average number of cells changed by a piece in its lifetime, the average number of pieces visiting a cell during a game.

To calculate the above mentioned statistics we might be able to construct the complete tree of all possible moves for some of the games. However, calculations on the basis of a complete tree may be misleading as we have earlier assumed that both the players are of equal intelligence and a complete tree is bound to have a considerable amount of foolish game play. A more realistic approach would be to calculate these quantities from actual game play. The amount of game play needed to have reliable statistics, especially if a population of games is to be evolved as is being done in this paper, prohibits human game play and our only choice is to have automated game play.

In the present work, a chromosome encodes the rules of a game. In other words, it is a complete game. The aim of the evolutionary process is to evolve a population of games and find a best one which is entertaining for the player. For this purpose we have assumed that better entertainment is based on four aspects described below. In three of these aspects we assume that both the players play each game with the same strategy (random controller). Hence both have the same chances of winning.

*3.3.1. Duration of the game:* In general, a game should not be too short or too long, as both are uninteresting. For example, if a game is such that it usually ends after a few moves (like Tic-Tac-Toe) then it would not appeal to adults. On the other hand, if a game usually continues for several hundred moves then the players may choose not to play it due to lack of enough time. The fitness function should be such that it discourages games which routinely end after a few moves as well as the games which take more than an upper threshold of moves.

The duration of play (D) of a game is calculated by playing the game n times and taking the average number of moves over these n games. For the games evolved in this paper, the maximum moves are fixed at 100 (50 for each player). If a game does not end in 100 moves then it is declared a draw. The average value of D is taken because if the game is played multiple times with a different strategy, (or even by the same strategy which has probabilistic components) then we do not get the same value of D every time. For averaging, the game is played n = 20 times in our experiments. Equation (1) shows the mathematical representation of D.

$$D = \frac{\sum_{k=0}^{n} L_K}{n} \qquad (1)$$

Where $L_K$ is the life of the game playing agent in game K. In order to reward games neither too short nor too long raw value of D is scaled in range 0-1. The boundaries for scaled value of D are shown in Fig. 2.

**Duration of Game D**

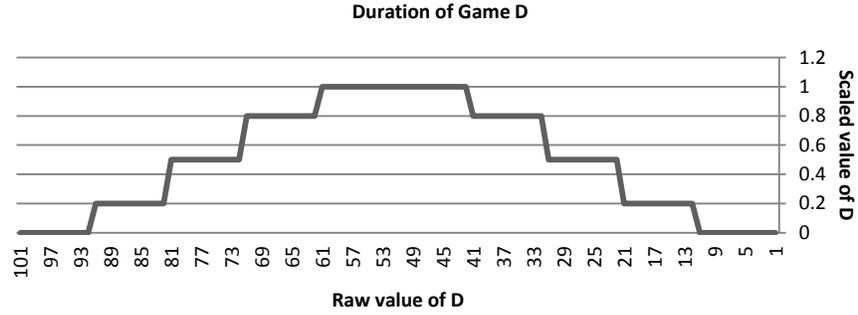

Fig. 2. Scaling ranges for raw value for duration of game.

For the raw duration of games 0 to 10 and 100 to 90 a scaled value of 0 is assigned, for ranges 11-20 and 81-90 a value of 0.2 is assigned, for 21-30 and 71-80 a value of 0.5 is used, 31-40 and 61-70 are converted to 0.8 and a range of 41-60 is assigned the highest value i.e. 1.

*3.3.2. Intelligence for playing the game:* A game is interesting if the rules of the game are such that the player having more intelligence should be able to win. In such a case, a win increases his motivation to play the game again. Too many draws or loses even after an intelligent game play are discouraging. There can be two reasons if a controller with superior intelligence does not win. Either the game is too simple or there are some inherent aspects which frustrate intelligent game play (e.g. very few pieces with limited step sizes making it impossible to corner the opponent).

The intelligence (I) is defined as the number of wins of an intelligent controller over a controller making random (but legal) moves. For this purpose the game is played *n* times (*n* = 20 times in our experiments). Higher number of wins against the random controller means that the game requires intelligence to be played and does not have too many frustrating dead ends. Intelligence I is calculated using equation (2).

$$I = \frac{\sum_{k=0}^{n} I_K}{n} \qquad (2)$$

Where, $I_K$ is 1 if intelligent controller wins the game otherwise its 0.

### 3.3.3. Game Liveliness:

*A. Dynamism exhibited by the pieces:*

This aspect assumes that a game whose rules encourage greater dynamism of movement in its pieces would be more entertaining than a game in which many pieces remain stuck in their cells for the entire duration of the game. The dynamism is captured by the following fitness function given in equation (3).

$$\text{Dyn} = \frac{\sum_{i=1}^{n} \left( \frac{\sum_{j=1}^{m}(c_j)/Li}{m} \right)}{n} \quad (3)$$

Where,

$C_i$ is the Number of cell changes made by piece i during a game

$L_i$ is life of the piece i

And m is the total number of pieces specified in a chromosome.

The dynamism is averaged by calculating it for 20 games for the same chromosome. This fitness function has a higher value if the pieces show a more dynamic behavior.

*B. Usability of the play area:* It is interesting to have the play area maximally utilized during the game. If most of the moving pieces remain in a certain region of the play area then the resulting game may seem strange. The usability is captured using equation (4).

$$U = \frac{\sum_{i=1}^{n} \left( \frac{\sum_{k=0}^{m}(c_k)}{|C_u|} \right)}{n} \quad (4)$$

where

$C_k$ usability counter value for a cell k.

$|C_u|$ is the total number of usable cells.

n is 20 as explained previously.

A usability counter is set up for each cell which increments when a piece arrives in the cell. The usability U is averaged by playing twenty different games for a chromosome. A cell which is never visited during a game will have a counter value of zero, thus contributing nothing to the usability formula. Furthermore, a cell which has a few visits would contribute less than a cell having large number of visits. This formula is another facet of the liveliness aspect.

*3.3.4. Combined fitness function:* The above four metrics are combined in the following manner. All chromosomes in a population are evaluated separately according to each of the four fitness functions. Then the population is sorted on "duration of game" and a rank based fitness is assigned to each chromosome. The best chromosome of the sorted population is assigned the highest fitness (in our case it is 20 because we have 10 parents and 10 offsprings), the second best chromosome is assigned the second best fitness (in our case 19), and so on. The population is again sorted on the basis of "intelligence" and a rank based fitness is assigned to each chromosome. Similarly, rank based fitness is assigned after sorting on "diversity" and "usability". The four rank based fitness values obtained for each chromosome are multiplied by corresponding weights and then added to get its final fitness.

$$FF = w_1 D + w_2 I + w_3 Dyn + w_4 U \quad (5)$$

where $w_1$, $w_2$, $w_3$, and $w_4$ are constants. In our experiments we keep the value of these constants fixed at 1. The multiplication with a corresponding weight allows us to control the relative influence of an aspect. The calculation of rank based fitness gets rid of the problem of one factor having higher possible values than another factor.

### 4. Software Agents for Playing the Games

Since evolutionary algorithm evolves a population of games and the fitness of each game has to be determined in each generation we may have a total of several thousands of such fitness evaluations. Fitness evaluation means playing the game several times, it is not possible to do so manually. We need software game playing agents. The more intelligent the agent the better will be the accuracy of fitness evaluation. We have developed two types of such agents.

- Random agent.
- Agent using Min-Max with rule based evaluation function.

### 4.1. Random Agent

As the name suggests the random game playing agent plays the game by randomly selecting a legal move at each step. The agent follows the following algorithm listed in Fig. 3:

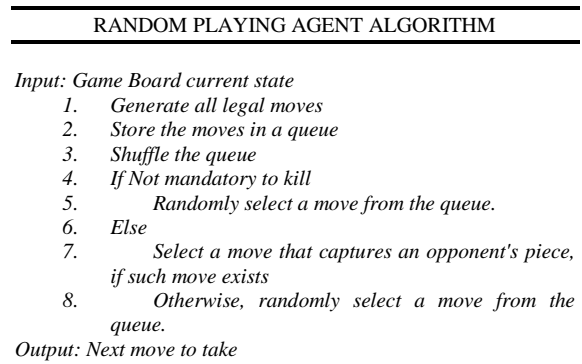

*Fig. 3. Algorithm for the random playing agent.*

The agent initially generates all the legal moves and stores them in a queue. The queue is shuffled once all the moves are saved in it. Shuffling is important, as we take an average of 20 games to calculate the individual metrics values. If the queue is not shuffled then each time the game is played it will use the same sequence of moves to play the game and fitness values will remain the same in each iteration of the game play. If the mandatory to capture bit is "on" in a chromosome which is being evaluated then the agent first tries to find a move that will capture an opponent's piece. If no such move is found it randomly selects a move from the queue.

### 4.2. Agent using Min-Max with rule based evaluation function

Intelligent agent generates all the possible one ply depth game boards using a min-max algorithm. Each of the resulting game board is evaluated using a rule based evaluation function and the one with the highest evaluation is selected as a next move.

Evaluation function assigns priorities (weights) to piece-type according to whether its disappearance would cause the game to end, flexibility of movement, and capturing logic. Once the priority of a piece is calculated we multiply each piece with its corresponding weight and calculate weighted summation for self and opponent. The board evaluation is the self-weighted summation minus opponents weighted summation. Fig. 4 lists the algorithm for the evaluation function we use.

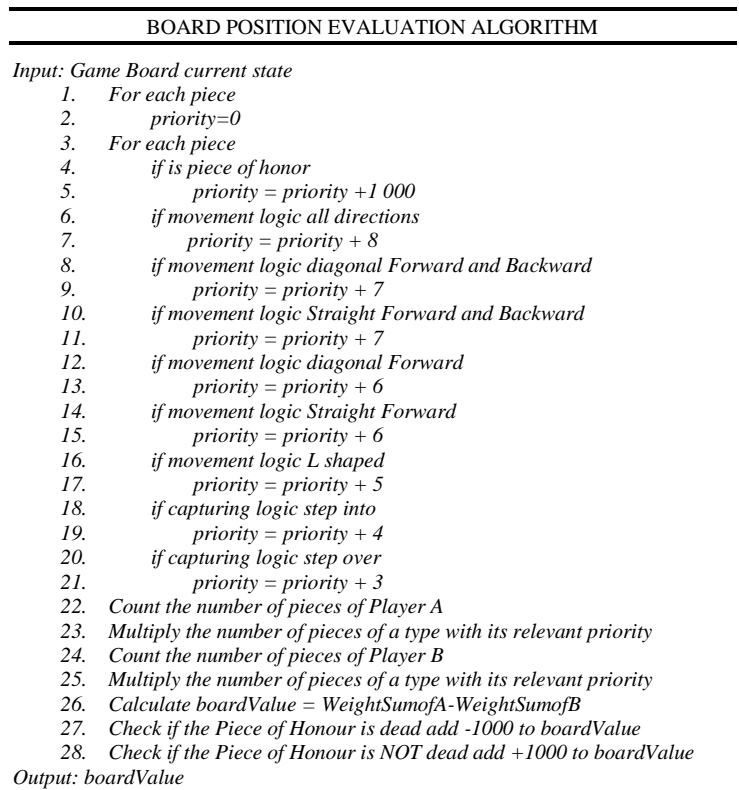

BOARD POSITION EVALUATION ALGORITHM

*Input: Game Board current state*
*1.    For each piece*
*2.        priority=0*
*3.    For each piece*
*4.        if is piece of honor*
*5.            priority = priority +1 000*
*6.        if movement logic all directions*
*7.            priority = priority + 8*
*8.        if movement logic diagonal Forward and Backward*
*9.            priority = priority + 7*
*10.       if movement logic Straight Forward and Backward*
*11.           priority = priority + 7*
*12.       if movement logic diagonal Forward*
*13.           priority = priority + 6*
*14.       if movement logic Straight Forward*
*15.           priority = priority + 6*
*16.       if movement logic L shaped*
*17.           priority = priority + 5*
*18.       if capturing logic step into*
*19.           priority = priority + 4*
*20.       if capturing logic step over*
*21.           priority = priority + 3*
*22.   Count the number of pieces of Player A*
*23.   Multiply the number of pieces of a type with its relevant priority*
*24.   Count the number of pieces of Player B*
*25.   Multiply the number of pieces of a type with its relevant priority*
*26.   Calculate boardValue = WeightSumofA-WeightSumofB*
*27.   Check if the Piece of Honour is dead add -1000 to boardValue*
*28.   Check if the Piece of Honour is NOT dead add +1000 to boardValue*
*Output: boardValue*

Fig. 4. Algorithm for evaluation of board positions.

Since we are using min-max of single ply hence we had to incorporate a mechanism in the evaluation function to overcome the randomness effect near the end of the game when pieces are few and may be far apart. In such cases the evaluation function listed in Fig. 4 gives same evaluation for all the board positions thus increasing the duration of game. To avoid such situation we restrict the agent to select the move which decreases distance between its own piece and one of an opponent's pieces, provided all next game board position have equal evaluation.

There can be other options to an intelligent controller; such as an artificial neural network (ANN) based controller. For an ANN based controller we would require training the connection weights and/or optimizing the ANN architecture. Thus ANN based controller will introduce computational overhead. Since the min-max algorithm is used with one ply depth, which overcomes the computational overhead.

## 5. Experiments

Different experiments carried out are explained in the sub-sections below. The methodology of the experimentations is such as we first conduct an experiment on the existing games of chess and checkers to calculate the individual values of our proposed metrics. The results of this experiment are shown in table 8-11 (appendix I). We will use these results to compare with the metrics values for the evolved games in section 5.1 of this paper. Afterwards we evolve new board based games using 1+1 Evolutionary Strategy (ES). Once we get a set of evolved games we first select minimum number of games, using equation (7), for analysis. In order to analyze and study the entertainment value contained in the evolved games we follow a threefold strategy i.e. by conducting a controller learnability experiment, a human user survey and lastly comparing the individual metrics values of the evolved games with the popular games of chess and checkers.

### 5.1. Evolution of New Games

In order to generate new and entertaining board based games we use 1+1 Evolutionary Strategy (ES). Initially a population of 10 chromosomes is randomly initialized with permissible values. The evolutionary algorithm is run for 100 iterations. Mutation is the only genetic operator used with a mutation probability of 30 percent. In each iteration of the ES one parent produce one child and a fitness difference is calculated between them. If it is greater than 4 (i.e. the child is at least half times better than its parent) child is promoted to the next population. We use the formula given in equation (6) to calculate the fitness difference.

$$\text{Fitness Difference} = \sum_{\text{for all metrics}} \left( 1 - \frac{\text{fitness}_p - \text{fitness}_c}{\text{fitness}_p} \right) \qquad (6)$$

Where,
$\text{fitness}_p$ is the fitness value of parent for current metrics
$fitness_c$ is the fitness value of child for current metrics

We keep an archive of 8 slots and in each iteration update it with the best 2 chromosomes based on each of the fitness metrics. Fig. 5 shows the metrics values of one family of chromosome in shape of a graph (Fig. 5), over a period of 100 iterations.

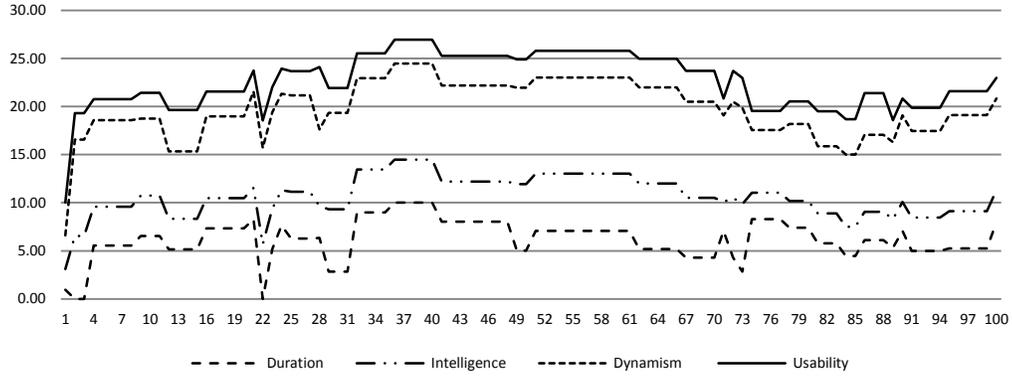

Fig. 5. Metrics values of a typical family of chromosome.

As we use 1+1 ES the best chromosome found in an iteration may get lost in iterations to come. For this purpose as mentioned previously we keep an archive of 8 for best 2 chromosomes against each of the metrics. As the evolutionary process progresses the number of changes, child beating its parent using fitness difference formula (equation (6)) decreases.

## 5.2. Games for Analysis

The evolutionary process gives 8 games evolved against the entertainment based on duration, intelligence, dynamism and usability. For further analysis of these games we select the set of most diverse games. Diversity (from each other) of these games is calculated using their fitness values and is listed in table 2 for one experiment.

Table 2. Fitness values of chromosomes in archive

|  | | Game No. | Duration | Dynamism | Intelligence | Usability |
|---|---|---|---|---|---|---|
| | Duration 1 | 1 | 0.89 | 0.08 | 1 | 21.05 |
| | Duration 2 | 2 | 0.85 | 0.07 | 0.7 | 16.78 |
| | Dynamism 1 | 3 | 0.02 | 0.18 | 1 | 22.07 |
| Archive | Dynamism 2 | 4 | 0.22 | 0.17 | 1 | 25.87 |
| | Intelligence 1 | 5 | 0 | 0.09 | 1 | 23.09 |
| | Intelligence 2 | 6 | 0 | 0.07 | 1 | 21.03 |
| | Usability 1 | 7 | 0.4 | 0.07 | 0.85 | 84.93 |
| | Usability 2 | 8 | 0.22 | 0.04 | 0.7 | 81 |

We calculate the diversity based on each of the four metrics for all pair of games using equation (7).

$$\text{Game diversity } = |\ \frac{\text{Game Column Fitness-Game Row Fitness}}{\text{Selected Metrics Maximum Value}}\ |\quad (7)$$

Selecting a threshold value of 0.6 table 3 shows the diversity count of evolved games. Diversity count indicates that a game is different from how many other games based on all the four metrics of entertainment.

Table 3. Diversity count of evolved games

| Game No. | Different from number of  games (for threshold ≥ 0.6) |
|----------|-------------------------------------------------------|
| 1 | 5 |
| 2 | 5 |
| 3 | 3 |
| 4 | 1 |
| 5 | 0 |
| 6 | 1 |
| 7 | 6 |
| 8 | 3 |

Based upon the above statistics game number 1, 2 and 7 seems to be the most diverse. We select these three for further analysis. From this point onwards we will refer to these as game 1, 2 and 3 respectively. Rules of these games are listed in Fig. 8 - Fig. 10 (appendix I). For the sake of simplicity we do not name the pieces rather we identify them by their type number which range from 1-6.

We also create a randomly initialized game that has not been passed through the evolutionary process for optimization of entertainment. This game is used to analyze the learnability of the game experiment covered in section 5.4 and also in the user survey, to compare with the evolved games. Rules of the random game are shown in Fig. 11 (appendix I).

### 5.3 Comparison of Evolved Games with Other Board Based Games

It would be quite informative to see performance comparisons of the proposed technique against other heuristic-type solvers used to create board based game. Other techniques like [21] may end up making a different board game thus two games may only be comparable, not the techniques. Having said this, game of chess has a history of 1500 years [34] and monopoly goes back to 1903. It would be more suitable to see the performance of evolved games and the game of chess and checkers using the of individual entertainment metrics (proposed earlier in this paper) as a gauge. Table 4 list the effect and values of each type of controller on the individual entertainment metrics. The results in table 4 suggest that the entertainment value of the popular games of chess and checkers are comparable to those evolved using the proposed entertainment metrics.

Table 4. Fitness values comparison of evolved games vs. chess and checkers

|  | Chess | Checkers | Game 1 | Game 2 | Game 3 |
|---|---|---|---|---|---|
| Duration | 0.30 | 0.60 | **0.89** | **0.85** | 0.40 |
| Dynamism | 0.10 | 0.30 | 0.08 | 0.07 | 0.07 |
| Intelligence | 1.00 | 0.50 | **1.00** | 0.70 | **0.85** |
| Usability | 3.30 | 3.10 | **21.05** | **16.78** | **84.93** |

The data listed in table 4 shows the average value of duration, dynamism, intelligence and usability recorded by playing each of the listed game 100 times. For the individual metrics of duration, intelligence and usability the evolved games have either performed better or at least equal to the games of chess and checkers.

### 5.4. Learnability of Evolved Games

The entertainment value of the evolved games needs to be verified against some criteria other than the proposed entertainment metrics. For this purpose we use the Schmidhuber's theory of artificial curiosity [7]. We need to see how quickly a player learns an evolved game. Games learned very quickly will be trivial for the player and thus not contributing anything towards entertainment. Those taking large amount of time to learn will be too difficult. Games between these two boundaries will fall in the range of entertaining games. To observe the learnability of the evolved games there are two options first is to ask a human to play a game multiple times and see how fast she/he learns and second is to do the same task using a software based controller.

We have used an ANN based controller. The architecture of the controller is the same architecture used by Chellapilla [30] for evolving an expert checkers player. There are total 5 layers in the ANN, input with 64 neurons, first hidden layer with 91 neurons; second hidden layer with 40 neurons third with 10 neurons and the output layer with 1 neuron. A hyperbolic tangent function is used in each neuron. The connection weights range from [-2, 2]. The training of the ANN is done using co-evolution. A set of genetic algorithm (GA) population is initialized that represent the weight of the ANN. Each individual of the population is played against randomly selected 5 others. Mutation is the only genetic operator used; we have kept the ANN and its training as close to Chellapilla [30] work as possible, except for the number of iteration for which the ANN is trained. We train the set of ANN until we get a set of weights that beats all others. Such individual will have its fitness equal to 1. The number of iterations that take to get such individual in the archive is called the learning duration or learnability of the game. Fig. 6 show the learnability of all the 4 games including the random game (referred to as game 4).

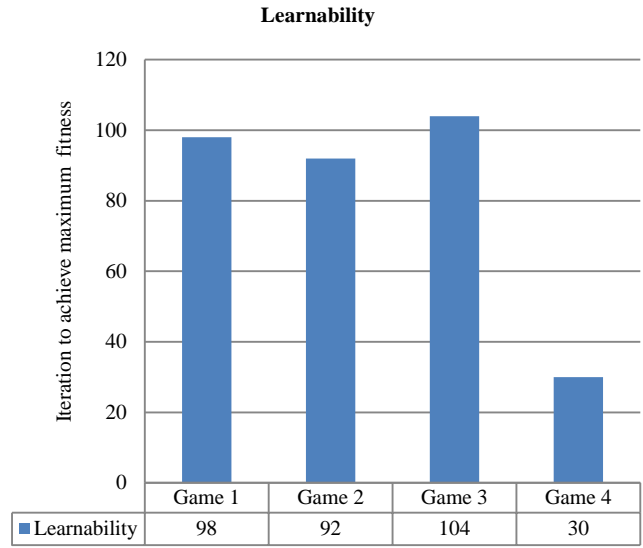

Fig. 6. Learnability of evolved and random game.

It takes about 80 to 110 iterations to get a chromosome representing ANN weights that achieve a fitness of 1 during co-evolution, for the evolved games. In case of the randomly initialized game (game 4) it takes only 30 iterations, thus showing game 4 is trivial and uninteresting. Thus we can conclude that game 1, 2 and 3 proof to be entertaining as an ANN based controller neither takes too short a time nor too long to learn these games.

## 5.5. User Survey on Entertainment Value of Evolved Games

To validate the results produced against human entertainment value we have to perform a human user survey. For this purpose we select a set of 10 subjects. Subjects are chosen such that they have at least some level of interest towards computer games. Each individual was given 4 games while s/he was supposed to play each game 3 times, the rules of the game were already explained to the subjects and also displayed on the software they used. This makes a total of 12 games to be played by each subject. The demographics of the subjects are summarized in table 5; listing subject age, gender and subject's likeness towards board games 1 being lowest and 10 being highest.

Table 5. Demographics of the subject for user survey

| Subject | Age | Gender | Like Computer Games [1-10] |
|---|---|---|---|
| 1 | 26 | M | 7 |
| 2 | 22 | F | 5 |
| 3 | 20 | M | 5 |
| 4 | 27 | M | 5 |
| 5 | 24 | F | 6 |
| 6 | 38 | M | 8 |
| 7 | 20 | M | 8 |
| 8 | 33 | M | 7 |
| 9 | 21 | M | 7 |
| 10 | 23 | F | 4 |

The four games given to the user are marked as Game 1, Game 2, Game 3 and Game 4. Rules are shown in Fig. 8- Fig. 11. Game 4 is the randomly initialized one (but will legal values) whereas remaining three games are evolved for entertainment. Each subject was asked to rank the game they play as 1- liked, 2-disliked and 3-neutral.

The results of the human user survey are shown in table 6. For visual purposes we use for √ liked, × for disliked and ~ for neutral.

Table 6. Human user survey results

| Subject | Game 1 Run 1 | Run 2 | Run 3 | Game 2 Run 1 | Run 2 | Run 3 | Game 3 Run 1 | Run 2 | Run 3 | Game 4 Run 1 | Run 2 | Run 3 |
|---|---|---|---|---|---|---|---|---|---|---|---|---|
| 1 | ~ | √ | √ | ~ | √ | √ | ~ | √ | √ | ~ | √ | × |
| 2 | ~ | ~ | √ | ~ | √ | √ | ~ | √ | √ | ~ | × | × |
| 3 | × | √ | √ | ~ | × | × | √ | × | × | × | × | × |
| 4 | ~ | √ | √ | √ | √ | √ | √ | √ | √ | √ | √ | √ |
| 5 | ~ | √ | √ | √ | √ | √ | ~ | √ | √ | ~ | √ | × |
| 6 | ~ | √ | √ | ~ | √ | √ | √ | √ | √ | ~ | √ | × |
| 7 | ~ | ~ | √ | ~ | √ | √ | √ | √ | √ | ~ | × | × |
| 8 | × | √ | √ | √ | √ | √ | ~ | √ | √ | × | × | × |
| 9 | √ | √ | √ | × | √ | √ | ~ | √ | √ | × | × | × |
| 10 | √ | √ | √ | × | √ | √ | ~ | √ | √ | ~ | ~ | ~ |

For the purpose of collecting statistics from the human user survey we mark a game liked by the subject if during any run he has liked the game and for rest he has either liked or been neutral. If in any run he has disliked the game we mark the game to be disliked. Fig. 7 shows the statistics based upon this scheme. About 70- 90 percent subjects have liked the evolved games and found these entertaining, whereas only 10 percent say that the random game (game 4) was entertaining.

**Game liked %**

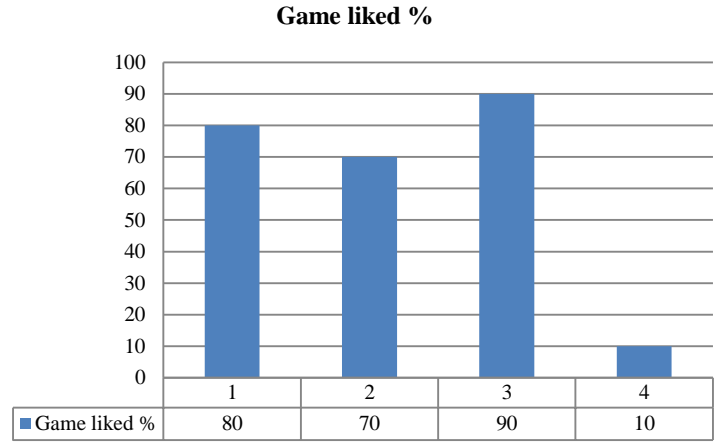

| | 1 | 2 | 3 | 4 |
|---|---|---|---|---|
| ■ Game liked % | 80 | 70 | 90 | 10 |

Fig. 7: Statistics of the human user survey

### 5.6. Hypothesis testing

For the purpose of testing whether or not the given hypothesis, that the evolved games are entertaining, is true, we conduct a hypothesis testing. For this purpose we first formulate the null hypothesis which is as follows:

H0: The correlation between the human value of entertainment and the entertainment contained in the evolved games, using combined fitness function, is a result of randomness.

The alternate hypothesis Ha is as

Ha: The evolved game using combined fitness function is entertaining.

To form a statistical test, we calculate a correlation coefficient c (z') using equation (8). Such that, given two games A and B, it can be established that A is more entertaining than B or otherwise, calculated as follows.

$$c(z^{'}) = \sum_{n=1}^{N} (\frac{z_n}{N}) \qquad (8)$$

N is the number of samples in the experiment.

$$Zn = \begin{array}{l} \textit{1, if subject finds game entertaining} \\ \textit{-1, if subject do not finds game entertaining} \\ \textit{0, otherwise} \end{array} \qquad (9)$$

We select the cut-off point, to reject null hypothesis, to be $\acute{a} = 0.17$. Table 7 shows the subject's answer to the question "Is game in row is more entertaining than game 4(random game)?" For visual purposes we use for √ yes, × for no and ~ for neutral. This data is collected after the user survey is completed from the same set of subjects.

Table 7. Answer to the question " is game in row is more entertaining than game 4(random game)?"

|  | Game 4 is entertaining | | | | |
|---|---|---|---|---|---|
| Game 1 | √ | √ | √ | × | √ |
|  | √ | √ | √ | √ | √ |
| Game 2 | √ | √ | √ | × | √ |
|  | √ | √ | √ | √ | ~ |
| Game 3 | √ | √ | √ | × | √ |
|  | √ | √ | √ | √ | √ |

Table 8. P-values for each pair of game

|  | C | $P(C \geq c)$ |
|---|---|---|
| Game 1-4 | 0.9 | 0.165282466 |
| Game 2-4 | 0.8 | 0.088439824 |
| Game 3-4 | 0.9 | 0.165282466 |

All the p-values listed in table 8 are below the cutoff point thus we reject the null hypothesis $H_0$, as a result alternate hypothesis $H_a$ is true.

## 6. Conclusion

The idea of evolution of game rules to produce new and entertaining games is intriguing. In this work we have presented automatic search of entertaining and new board based games. For this purpose we initially define a search space that is extracted using the popular board based games of chess and checkers. This search space is provided in shape of population of chromosomes to the evolutionary algorithm. To guide the evolutionary algorithm towards entertaining games we propose a set of metrics extracted based on different theories of entertainment. The metrics include duration of game, intelligence required to play the game, dynamism exhibited by the pieces and the usability of the play area. The results of our experiments are verified against the human user's entertainment value by conducting a human user survey and an experiment on controller learning ability. Results suggest that the evolved games using the fitness function (entertainment metrics) are interesting and much better than randomly generated games. The idea of measuring entertainment and automatic generation of entertaining games can be extended to other genres of game like platform games and real time strategy games. It will be interesting to see the effect of different types of controllers that play the game to evaluate its fitness on the entertainment value of the evolved games

## Appendix I

| Piece No | Movement Logic | Step Size | Capturing Logic | Conversion Logic |
|----------|----------------|-----------|-----------------|------------------|
| 1 | L | Multiple | Step Into | 6 |
| 2 | Diagonal Forward & Backward | Single | Step Over | 5 |
| 3 | All Directions | Multiple | Step Into | Nil |
| 4 | Straight Forward | Multiple | Step Into | 1 |
| 5 | Straight Forward | Multiple | Step Over | 2 |
| 6 | All Directions | Multiple | Step Over | 3 |

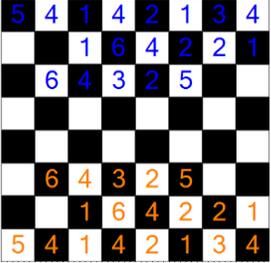

Piece of Honor     5

Mandatory to Capture     No

Fig. 8: Rules of game 1 and pieces board positions

| Piece No | Movement Logic | Step Size | Capturing Logic | Conversion Logic |
|----------|----------------|-----------|-----------------|------------------|
| 1 | L | Single | Step Over | 1 |
| 2 | Diagonal Forward | Single | Step Into | 3 |
| 3 | Diagonal Forward | Single | Step Over | Nil |
| 4 | All Directions | Multiple | Step Over | Nil |
| 5 | Straight Forward | Multiple | Step Into | 2 |
| 6 | All Directions | Single | Step Into | 1 |

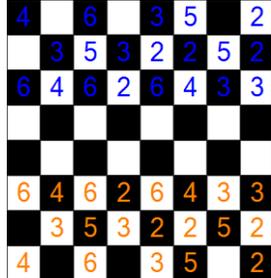

Piece of Honor     5

Mandatory to Capture     Yes

Fig 9. Rules of game 2 and pieces board positions

| Piece No | Movement Logic | Step Size | Capturing Logic | Conversion Logic |
|----------|----------------|-----------|-----------------|------------------|
| 1 | L | Multiple | Step Into | 5 |
| 2 | L | Single | Step Into | 2 |
| 3 | Straight Forward | Multiple | Step Over | 3 |
| 4 | Straight Forward | Single | Step Over | 5 |
| 5 | Diagonal Forward & Backward | Single | Step Over | 3 |
| 6 | Diagonal Forward | Multiple | Step Over | 3 |

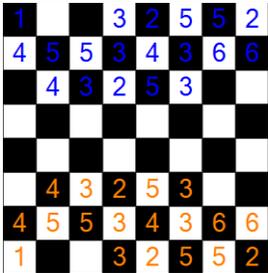

Piece of Honor     Nil

Mandatory to Capture     Yes

Fig. 10. Rules of game 3 and pieces board positions

| Piece No | Movement Logic | Step Size | Capturing Logic | Conversion Logic |
|----------|----------------|-----------|-----------------|------------------|
| 1 | All Directions | Multiple | Step Into | Nil |
| 2 | All Directions | Single | Step Over | 1 |
| 3 | Straight Forward & Backward | Multiple | Step Over | 5 |
| 4 | Diagonal Forward & Backward | Multiple | Step Over | 2 |
| 5 | Straight Forward | Multiple | Step Into | 6 |
| 6 | Straight Forward & Backward | Single | Step Into | 6 |

Piece of Honor        5

Mandatory to Capture   No

Fig. 11. Rules of random game and pieces board positions